\documentclass{bmvc2k}

\usepackage{graphicx}
\usepackage{amsmath,amssymb} 
\usepackage{color}
\usepackage{array}
\usepackage{booktabs}
\usepackage{color}
\usepackage{algorithm}
\usepackage{algpseudocode}
\usepackage{wrapfig}

\title{PCN: Part and Context Information \\
     for Pedestrian Detection with CNNs}

\addauthor{Shiguang Wang}{xiaohu_wyyx@163.com}{1}
\addauthor{Jian Cheng}{chengjian@uestc.edu.cn}{1}
\addauthor{Haijun Liu}{haijun_liu@126.com}{1}
\addauthor{Ming Tang}{s1646687@sms.ed.ac.uk}{2}
\addinstitution{
School of Electronic Engineering, \\
University of Electronic Science and Technology of China
}
\addinstitution{
 Science$\&$Engineering,\\
 The University of Edinburgh
}

\runninghead{Wang et al.}{PCN for Pedestrian Detection}

\begin{document}

\maketitle

\begin{abstract}
Pedestrian detection has achieved great improvements in recent years, while complex occlusion handling is still one of the most important problems. To take advantage of the body parts and context information for pedestrian detection, we propose the part and context network (PCN) in this work. PCN specially utilizes two branches which
detect the pedestrians through body parts semantic and context information, respectively. In the Part Branch, the semantic information of body parts can communicate with each other via recurrent neural networks. In the Context Branch, we adopt a local competition mechanism for adaptive context scale selection. By combining the outputs of all branches, we develop a strong complementary pedestrian detector with a lower miss rate and better localization accuracy, especially for occlusion pedestrian.
Comprehensive evaluations on two challenging pedestrian detection datasets (i.e. Caltech and INRIA) well demonstrated the effectiveness of the proposed PCN.
\end{abstract}

\section{Introduction}
\label{sec:intro}
Pedestrian detection, a canonical sub-problem in computer vision, has been extensively studied in recent years
\cite{Zhang2016Is,Luo2014Switchable,Tian2015Pedestrian,
Hosang2015Taking,Tian2015Deep,Cai2015Learning,Zhang2016How}. It has many applications such as video surveillance~\cite{bilal2016low} and intelligent vehicles~\cite{broggi2016intelligent}. Although pedestrian detection has achieved steady improvements over the last decade, complex occlusion is still one of the obstacles. According to the statistic analysis in \cite{Doll2012Pedestrian}, over 70$\%$ of the pedestrians are occluded at least in one video frame. For example, the current best-performing detector
RPN+BF~\cite{Zhang2016Is} attained 7.7$\%$ of the average miss rate on Caltech \cite{Doll2012Pedestrian} under the none occlusion setting. However, when heavy occlusions are present, its performance is much worse than several state-of-the-art algorithms. 
The detailed results are shown in Table~\ref{caltech result table}.

\begin{wrapfigure}{r}{6.5cm}
\centering
\includegraphics[width=6.5cm]{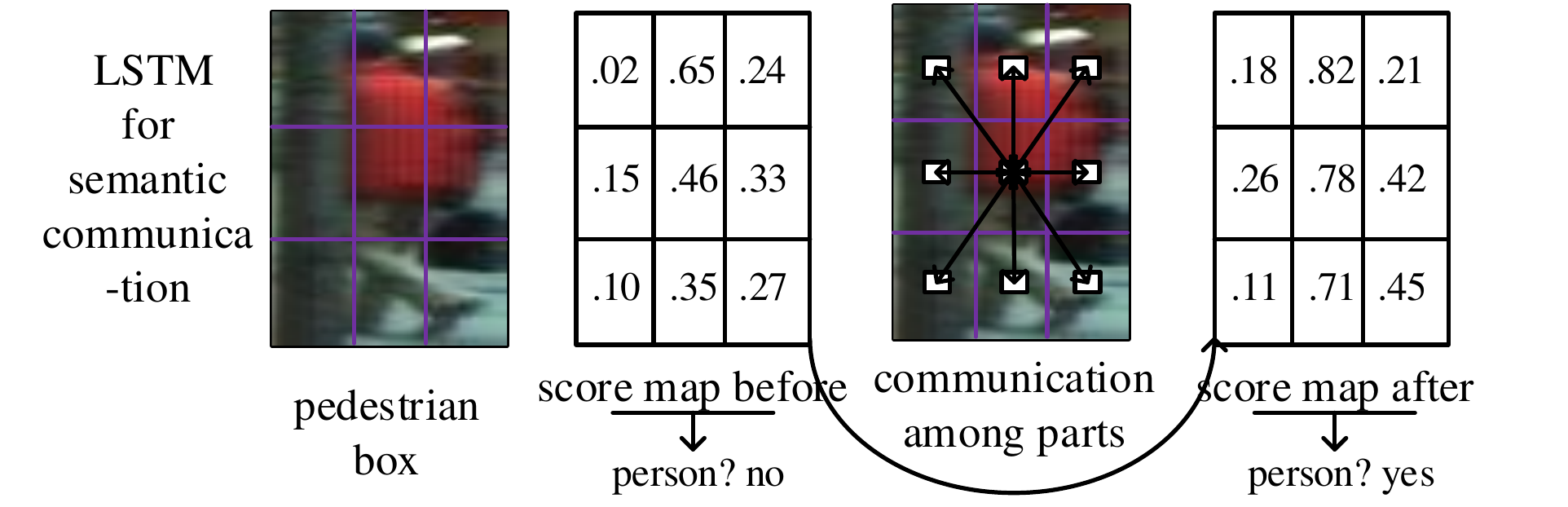}
\caption{Motivation of our part branch. It's hard to classify the pedestrian box because of heavy occlusion. With communication among parts, the visible parts and invisible parts could pass message bi-directionally and complement each other to support the existence of pedestrian. Thus, the refined body parts score map is prone to classify the pedestrian box.}
\label{fig:part-lstm-illu}
\end{wrapfigure}

Recently, DeepParts~\cite{Tian2015Deep} was specially designed to cope with the occlusion cases in pedestrian detection, by constructing an extensive part pool for automatic parts selection. The well-designed parts are selected driven by data, requiring less prior knowledge of the occlusion types than~\cite{Mathias2013Handling,Wojek2011Monocular}. However, the semantic parts cannot communicate with each other, which is very important for occluded pedestrian handling. As illustrated in Fig.~\ref{fig:part-lstm-illu}, when left hand is occluded, the corresponding part score may be very low. Consequently, the summed score will also be low. The final result is failed to recognize the occluded pedestrian. However, with semantic communication, the visible parts such as head and middle body can pass messages to support the existence of left hand.
Thus, the body part scores could be refined by this semantic information communication and we are more likely to recognize the occluded pedestrian successfully.

Inspired by ION~\cite{Bell2016Inside}, we introduce LSTM~\cite{Surhone2010Long} for  semantic parts communication due to its extraordinary ability for learning long-term dependences. As shown in Fig.~\ref{fig:part-lstm-illu}, the pedestrian box is divided into several part grids. Each part grid corresponds to a detection score. With LSTM, those body parts could memorize the semantic information and communicate with other parts to refine the score map. Moreover, different body parts need different information from other semantic parts, therefore, we use the gates function in LSTM to control the message passing.

On the other hand, a lot of researches \cite{Bell2016Inside, Gidaris2015Object,cai2016unified} stressed that context information plays an important role in object detection. However, they usually only extracted the context information in a single scale, where the scale is set by handcraft \cite{Gidaris2015Object,cai2016unified}.
Considering that different pedestrian instance may need different context information, we propose to use context regions with different scales to extract the context information. To adaptively select the context information in different scales, a local competition operation (maxout~\cite{Goodfellow2013Maxout}) is introduced into our model.
For example, the small size pedestrian may need more context information due to low resolution, whilst the heavy-occluded pedestrian may need more context information than the no-occluded pedestrian.

\noindent \textbf{Contributions.} To sum up, this work makes the following contributions. (1) we propose a novel part and context network (PCN) for pedestrian detection, which incorporates an original branch, a part branch and a context branch into an unified architecture. (2) To the best of our knowledge, we are the first to introduce the LSTM module into pedestrian detection framework for body parts semantic information communication. (3) The context branch can adaptively handle multi-scale context information in a data driven manner via a local competition mechanism. (4) The results on Caltech and INRIA datasets demonstrated the effectiveness of our proposed model PCN for pedestrian detection with a lower miss rate and a better localization accuracy, especially for those occluded pedestrians.

\section{Related Work}
In this section, we mainly review related works in three aspects.

\textbf{Part-Based Pedestrian Detectors}. Part-based approaches for pedestrian detection can be classified into two categories: supervised manner and unsupervised manner. Mohan et al.~\cite{Mohan2001Example} and Krystian et al.~\cite{Mikolajczyk2004Human} firstly trained part detectors in a fully supervised manner and then combined their outputs to fit a geometric model. Another category of part-based models focus on unsupervised part mining, which does not require part labels. Wang et al. \cite{Wang2012A} introduced Deformable Part Model (DPM) to handle the pose variations by learning a mixture of local templates for each body part. Lin et al. \cite{Lin2015Discriminatively} proposed a promising framework by incorporating DPM into And-Or graph. However, DPM~\cite{Wang2012A} needs to handle complex configurations. Recent work DeepParts ~\cite{Tian2015Deep} constructed an extensive part pool for automatic parts selection, but its performance was restricted by the handcrafted part pool.

\textbf{Multi-Context Features}. Multi-context or multi-region features are useful in object detection~\cite{Bell2016Inside, Gidaris2015Object, Ouyang2017Learning, cai2016unified}. Gidaris et al.~\cite{Gidaris2015Object} firstly introduced multi-region into deep convolutional neural network and explored their role in detection. Bell et al~\cite{Bell2016Inside} used spatial recurrent neural networks to integrate muti-context information of RoI. Ouyang et al.~\cite{Ouyang2017Learning} applied RoI-Pooling from image features using
different context regions and resolutions.

\textbf{Deep Models}. Deep learning methods are widely used in pedestrian detection. For instance, Hosang~\cite{Hosang2015Taking} demonstrated the effectiveness of the R-CNN pipeline~\cite{girshick2014rich} in pedestrian detection. Cai et al. learned complexity-aware cascades for deep pedestrian detection~\cite{Cai2015Learning} and also proposed a unified multi-scale deep CNN for fast object detection~\cite{cai2016unified}. Zhang et al.~\cite{Zhang2016Is} used a Region Proposal Network (RPN)~\cite{Ren2016Faster} to
compute pedestrian candidates and a cascaded Boosted Forest~\cite{appel2013quickly} to perform sample re-weighting for candidates classification. Du et al.~\cite{DBLP:journals/corr/DuELD16} introduced a soft-rejection based network fusion method to fuse the soft metrics from all networks together to generate the final confidence scores, achieving state-of-the-art results on Caltech pedestrian dataset~\cite{Doll2012Pedestrian}.

\section{Our Approach}
To take advantage of the body parts and context information for pedestrian detection, we propose the part and context network (PCN). PCN specially utilizes two sub-networks which detect the pedestrians through body parts semantic information and context information, respectively.
\subsection{Architecture}

\begin{figure}
\centering
\includegraphics[width=11.6cm]{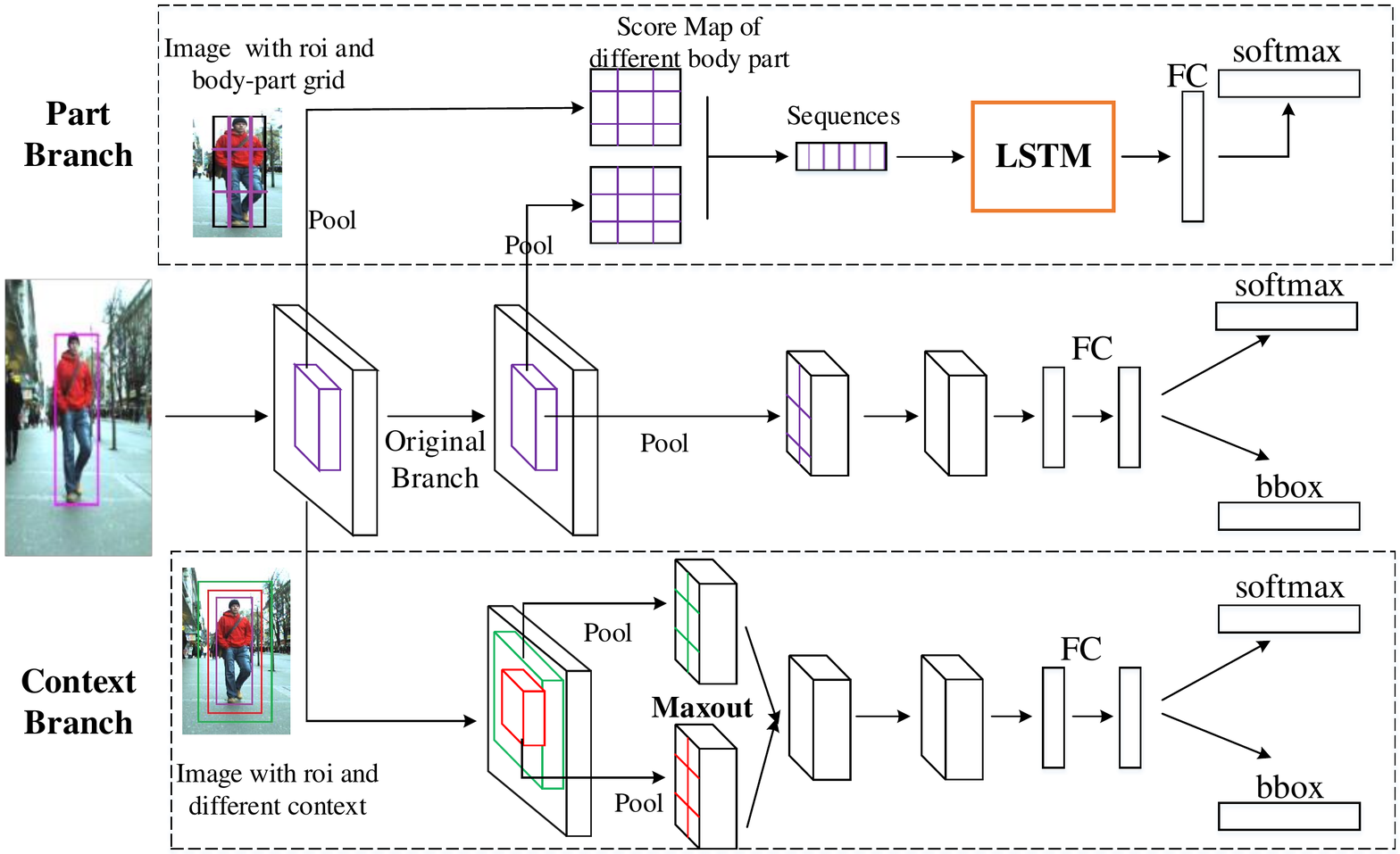}
\caption{Architecture of our model. The unified network consists of 3 subnetworks: a part branch using LSTM for semantic information communication, an original branch using original box from RPN, a context branch introducing maxout operation for region scale selection.}
\label{fig:framework}
\end{figure}

The detailed architecture of our pedestrian detector is shown in Fig.~\ref{fig:framework}. Based on the framework of Faster RCNN~\cite{Ren2016Faster}, our network detects objects through 3 detection branches: the original branch, the part branch and the context branch. Each branch has different focus.
The original branch uses original box from RPN~\cite{Ren2016Faster} so that it can focus on the full body of pedestrian.
The part branch is designed to make full use of semantic parts information for precise classification, specially for pedestrians with heavy occlusion. Considering that pedestrian instances may have different occlusion states,
part score maps related to pedestrian box are modelled as
sequence problem to communicate with each other. In the
part branch, pedestrian box is divided into several part grids (e.g., 3$\times$3) after RoI-Pooling and several convolution layers, where each part grid is associated with a detection score. Those semantic scores are applied for pedestrian parts communication using LSTM.
Since multi-region, multi-context features were found to be effective for object detection~\cite{Bell2016Inside, Gidaris2015Object, Ouyang2017Learning}, a context branch is introduced to obtain context features with diversity among multiple scales. Under the motivation that different pedestrian instances may need different context information, we adopt a local competition mechanism (maxout) among multiple scales to increase the adaptive ability of context selection.
The results produced by all detection branches are weighted sum as the final detection scores.
In our implementation, ``$ \hat{a} $ trous'' ~\cite{Chen2014Semantic} trick is used  to increase the feature map resolution.

\subsection{LSTM for part semantic information communication}

\begin{figure}
\centering
\includegraphics[width=12.6cm]{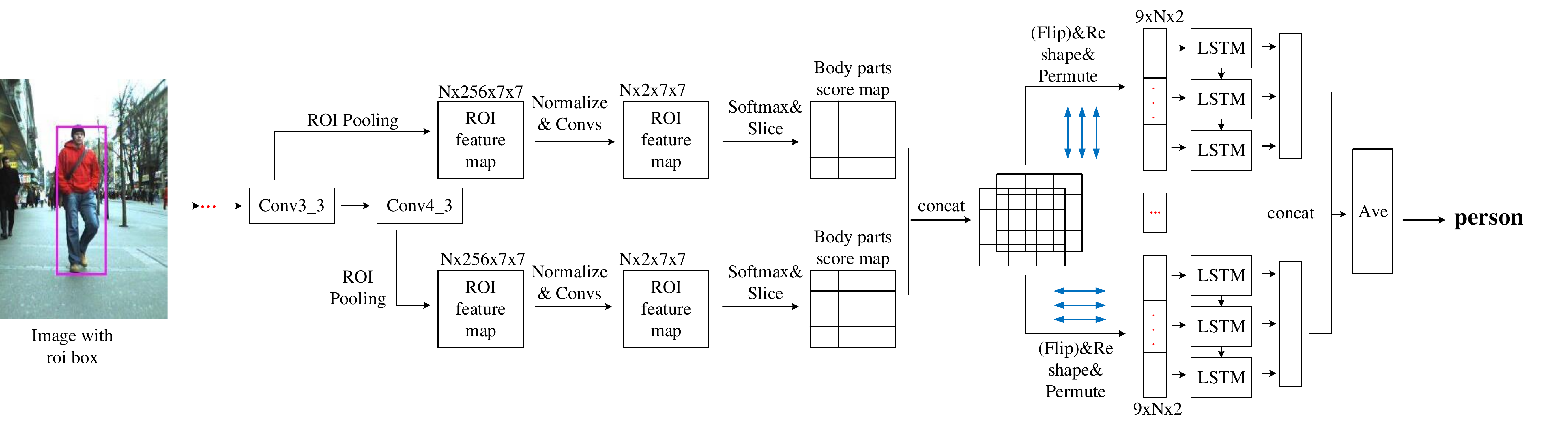}
\caption{LSTM for part semantic communication. In a single pass, we extract RoI feature from $conv3\_3$ and $conv4\_3$. Each descriptor is $L2$-normalized. We get parts score maps with size of 3 $\times$ 3 through several conv layers. The body parts score maps after softmax are concatenated, (flipped or not), reshaped, permuted and then sent to LSTM networks. LSTM moves along four directions: left, right, top and bottom. Refined scores outputted from LSTM are averaged to get the final scores of Part Branch.}
\label{part framework}
\end{figure}

Our architecture for part semantic features communication in PCN is shown in detail in Fig.~\ref{part framework}. To get RoI features with different resolution, we place RoI-Pooling layer on top of different conv layers. Features of $conv3\_3$ and $conv4\_3$ achieved best performance in our experiments.
$L2$-normalization is used following \cite{Liu2015ParseNet} to make training more stable. To get a parts score map of the pedestrian instance, the RoI feature maps are passed through several convolution layers to reduce channel dimension and resolution. Each branch gets a parts score map with resolution of K$\times$K (K=3 in Fig.~\ref{part framework}) after the softmax layer. The score maps have strong semantic information of pedestrian, for example the top-left grid indicates the left-head-shoulder part. These parts score maps are concatenated for communication in the next step.

To model coherence between semantic parts, we introduce LSTM for encoding their relationships due to its long short term memory ability to model sequential data. We first permute, flip, and reshape the parts score map to generate 4 sequences with different orders. Then, the LSTM moves on parts score map along four directions: left, right, top and bottom to encode parts and make semantic communication among them.
As demonstrated in Fig.~\ref{fig:part-lstm-illu} in Section~\ref{sec:intro} , when left hand is occluded, the corresponding part score may be very low. However, the visible part such as head and middle body can pass message to support the existence of the left hand. Consequently, the body part scores could be refined by this semantic information communication. The gate functions in LSTM are used to control the message passing.

\subsection{Maxout for adaptive context selection}

\begin{figure}
\centering
\includegraphics[width=11.6cm]{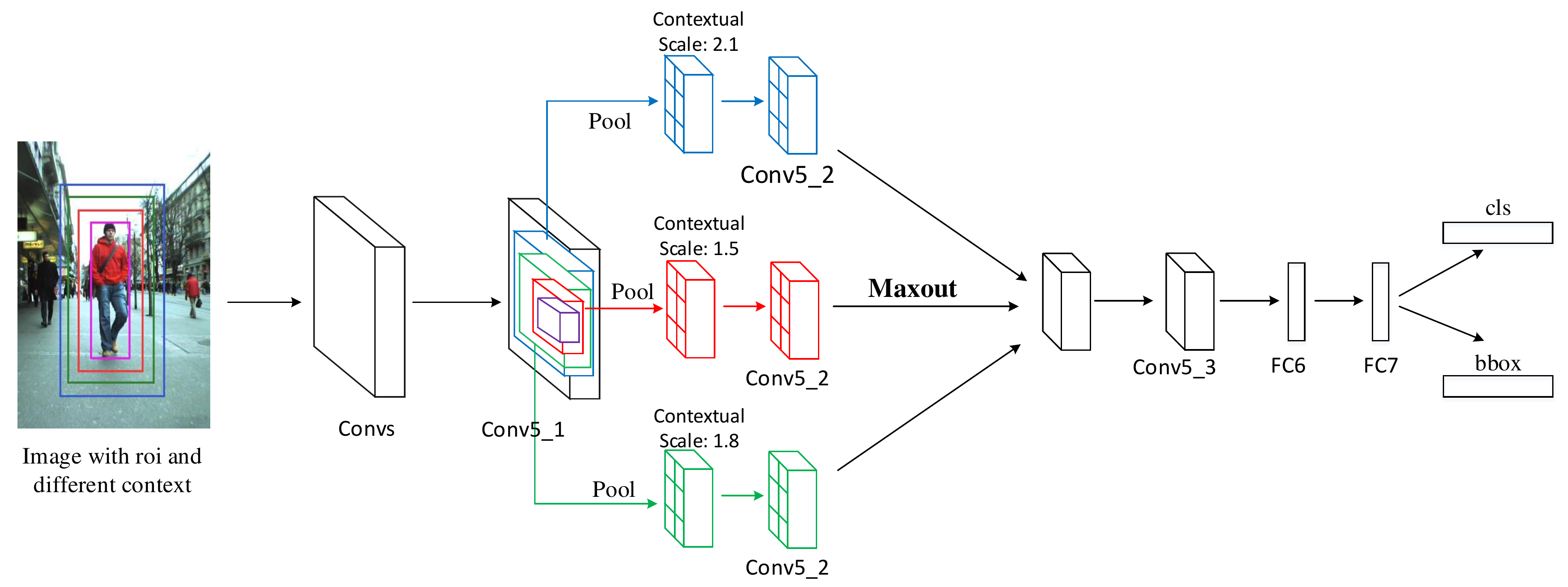}
\caption{Maxout for adaptive context selection. As shown above, the purple box indicates the original box of pedestrian, the red, green and blue box indicate the original context box with a scale factor of 1.5, 1.8 and 2.0, respectively. After a convolution layer, each context box is sent to maxout for competition and adaptive context selection. Class confidence and bounding box are outputted as the result.}
\label{context framework}
\end{figure}

Our architecture for adaptive context selection in PCN is detailedly shown in Fig.~\ref{context framework}.
Multi-region, multi-context features were found to
be effective in
\cite{Bell2016Inside, Gidaris2015Object,cai2016unified} . However, they only extracted the context information in a single scale, where the scale is set by handcraft \cite{Gidaris2015Object,cai2016unified}.
Considering that different pedestrian instance may need different context information, we propose to use context regions with different scales to extract the context information. For example, small size pedestrian may need more context information due to low resolution, and heavy-occluded pedestrian may need more context information than no-occluded pedestrian. Similar philosophy was also figured out in generic object detection~\cite{Ouyang2017Learning}, which used a chaining structure for multi-context mining. Our context convolutional feature maps are extracted from multiple discrete scales, known as multi-context. In Fig.~\ref{context framework}, 3 kinds of region scale are enumerated for generating context box. Suppose the RoI has width $ W $, height $H$, and $S$ is the context scale factor of the RoI. Thus, the context region has width $ W \times S $ and height $ H \times S $ with the same center as the original RoI, where $S=1.5,1.8,2.1$ respectively for each context branch in our implementation.

We incorporate a local competition operation (maxout) into PCN to improve the adaptive context selection capability for various context features sources.
Each sub-branch with different context scale passes through RoI-Pooling layer to get a fixed-resolution ($m \times m$) object feature map. These feature maps of different data source are selected by maxout with a data driven manner.
Maxout~\cite{Goodfellow2013Maxout} (element-wise max) is a widely considered operation for merging two or multiple competing sources. Here, we follow the well-known structure of NoCs~\cite{Ren2015Object}. When the maxout operation is used, the three feature maps (for the three context scales) are merged into a single feature map with the same dimensionality. The convolution layers before each RoI-Pooling layer share their weights. Thereby, the total number of weights is almost unchanged when using maxout.

\subsection{Training Procedure}
Since these branches differ a lot from feature extraction and expression, which could lead to some instability during learning, we train our model by multiple stages.

Stage1: Jointly optimizing original branch and context branch. Original branch and context branch could be jointly
optimized mainly because they have the similar feature expression (different from part branch). Therefore, the gradients from the 2 branches could be merged to the trunk in a stable way. The parameters $W$ of original branch and context branch are learned from a set of training samples. Suppose $X_i$ is a training image patch, and $Y_i=(y_i, t_i)$ the combination of its class label $y_i \in \{0,1,2,..K\}$ and bounding box coordinates $t_i=(t_i^x,t_i^y,t_i^w,t_i^h)$, we optimize $W$ with a multi-task loss:
\begin{equation}
L(W) = \sum_{m=1}^{M}\alpha_{m}l^{m}(X_i,Y_i|W),
\end{equation}
where $M$ is the number of detection branches, and $\alpha_m$ is the weight of loss $l^m$. $l^m$ is a multi-task loss combining the cross-entropy loss and the smoothed bounding box regression \cite{Ren2016Faster,Girshick2015Fast}. The optimal parameters $W^*=arg\min_{W}L(W)$ are learned by stochastic gradient descent.

Stage2: Pre-training the part branch to get parts score maps by the cross-entropy loss $L_{cls}(p(X),y)=-\log p_y(X)$, where $p(X)=(p_0(X),...,p_K(X))$ is the probability distribution over classes. Part branch is pre-trained so that parts can focus on information communication in Stage3. The model trained in Stage1 is fixed in Stage2 to save compute cost.

Stage3: Training the part branch for parts semantic information communication using LSTM. Parameters trained in stage1 are fixed to save compute cost, while parameters trained in Stage2 are finetuned in Stage3.

\section{Experiments}
We evaluated the effectiveness of the proposed PCN
on 2 popular pedestrian detection datasets including Caltech~\cite{Doll2012Pedestrian}, INRIA~\cite{Dalal2005Histograms}. More experimental analyses on the effectiveness of each component
in our network are further given on the challenging Caltech dataset~\cite{Doll2012Pedestrian}.
\subsection{Datasets}
\begin{table}
\footnotesize
  \centering
  \begin{tabular}{c|c||c|c}
    \toprule[2pt]
    Settings & Description & Settings & Description
    \\ \toprule[1pt]	
    Reasonable & 50+ pixels, Occ. none or partial & Occ.partial & 1-35$\%$ occluded \\
    All & 20+ pixels, Occ. none, partial or heavy & Occ.heavy & 35-80$\%$ occluded\\
    Occ.none & 0$\%$ occluded &  Over.75  & Reasonable,  using IOU=0.75\\

    \toprule[2pt]
  \end{tabular}
  \caption{Evaluation settings for Caltech pedestrian dataset~\cite{Doll2012Pedestrian}.}
  \label{evaluate settings table}
\end{table}
\noindent \textbf{Caltech}.
The Caltech dataset and its associated benchmark~\cite{Doll2012Pedestrian} are among the most popular pedestrian detection datasets. It consists of about 10 hours videos (640$\times$480) collected from a vehicle driving through regular urban traffic.
The annotation includes temporal correspondence between bounding boxes and detailed occlusion labels. We use dense sampling of the training data by 10 folds as adopted in~\cite{Hosang2015Taking}. 4024 images in the standard test set are used for evaluation. The miss rate (MR) is used as the performance evaluation metric~\cite{Doll2012Pedestrian}. The detailed evaluation settings following~\cite{Doll2012Pedestrian} are shown in Table~\ref{evaluate settings table}.

\noindent \textbf{INRIA}.
The INRIA pedestrian dataset~\cite{Dalal2005Histograms}, which is often used for verifying the generalization capability of models, is split into a training and a testing set. The training set consists of 614 positive images and 1,218 negative images. The testing set consists of 288 testing images. Our model is evaluated on the testing set by MR.

\subsection{Implementation Details}

\noindent For RPN, we used anchors of 9 different scales, starting from 40 pixels height with a scaling stride of 1.4$\times$. Other hyper-parameters of RPN followed~\cite{Zhang2016Is,Ren2016Faster,Girshick2015Fast}. When training, 70k iterations were run with an initial learning rate of 0.001, which decays 10 times after 50k iterations. Model was finetuned from 16-layers VGG-Net. When testing, the scale of the input image was set as 720 and 560 pixels on the shortest side on Caltech and INRIA datasets respectively, we just selected top 50 boxes for proposals.

\noindent For PCN, we used the hyper-parameters following~\cite{li2015scale} to finetune our model from 16-layers VGG-Net. The fourth max pooling layer was removed to produce larger feature maps in all branches. ``$ \hat{a} $ trous''~\cite{Chen2014Semantic} trick was used  to increase the feature map resolution and reduce stride. Based on these settings, the parameter optimized model was used as our basic model (RPN+FCNN$_{opt}$). When training, top 1000 boxes were selected as proposals, and 50k iterations were run with an initial learning rate of 0.001 for 3 stages, which decays 10 times after 40k iterations. When testing, with the top 50 boxes from RPN, we used nms with 0.5 after weight combined from different branches. Conv5's channels are reduced to 256, while fc6 and fc7 are reduced to 2048, 512 respectively to reduce computation and memory burden.

\subsection{Comparisions with State-of-the-art Methods}

\begin{table}
\footnotesize
  \centering
  \begin{tabular}{p{2.6cm}|c|c|c|c|c|c}
    \toprule[2pt]
    Methods & Reasonable & All & Occ.none & Occ.partial & Occ.heavy & Over.75  \\ \toprule[1pt]	\toprule[1pt]	
    SCF+AlexNet~\cite{Hosang2015Taking}   & 23.3 & 70.3  & 20.0 & 48.5 & 74.7 & 58.9 \\
    DeepParts~\cite{Tian2015Deep}    & 11.9 & 64.8 & 10.6 & 19.9 & 60.4 & 56.8 \\
    CompACT-Deep~\cite{Cai2015Learning} & 11.8 & 64.4 & 9.6 & 25.1 & 65.8 & 53.3\\
    SAF R-CNN~\cite{li2015scale}    & 9.7 & 62.6 & 7.7 & 24.8 & 64.4 & 41.0 \\
    MS-CNN~\cite{cai2016unified}      & 10.0 & \textbf{61.0} & 8.2 & 19.2 & 59.9 & 56.9\\
    RPN+BF~\cite{Zhang2016Is}       & 9.6 & 64.7 & 7.7 & 24.2 & 69.9 & 35.5 \\
    RPN+FRCNN$_{opt}$~\cite{Ren2016Faster}   & 12.1  & 65.4  & 10.5 & 24.0 & 64.9 & 48.5 \\
    PCN(Ours) & \textbf{8.4} & 61.8 & \textbf{7.0} & \textbf{16.4} & \textbf{56.7} & \textbf{34.8}  \\
    \toprule[2pt]
  \end{tabular}
  \caption{Detailed breakdown performance comparisons of our models and other state-of-the-art models on the 6 evaluation settings.}
  \label{caltech result table}
\end{table}

\begin{figure}
\begin{tabular}{cc}
\bmvaHangBox{\fbox{\includegraphics[width=5.6cm]{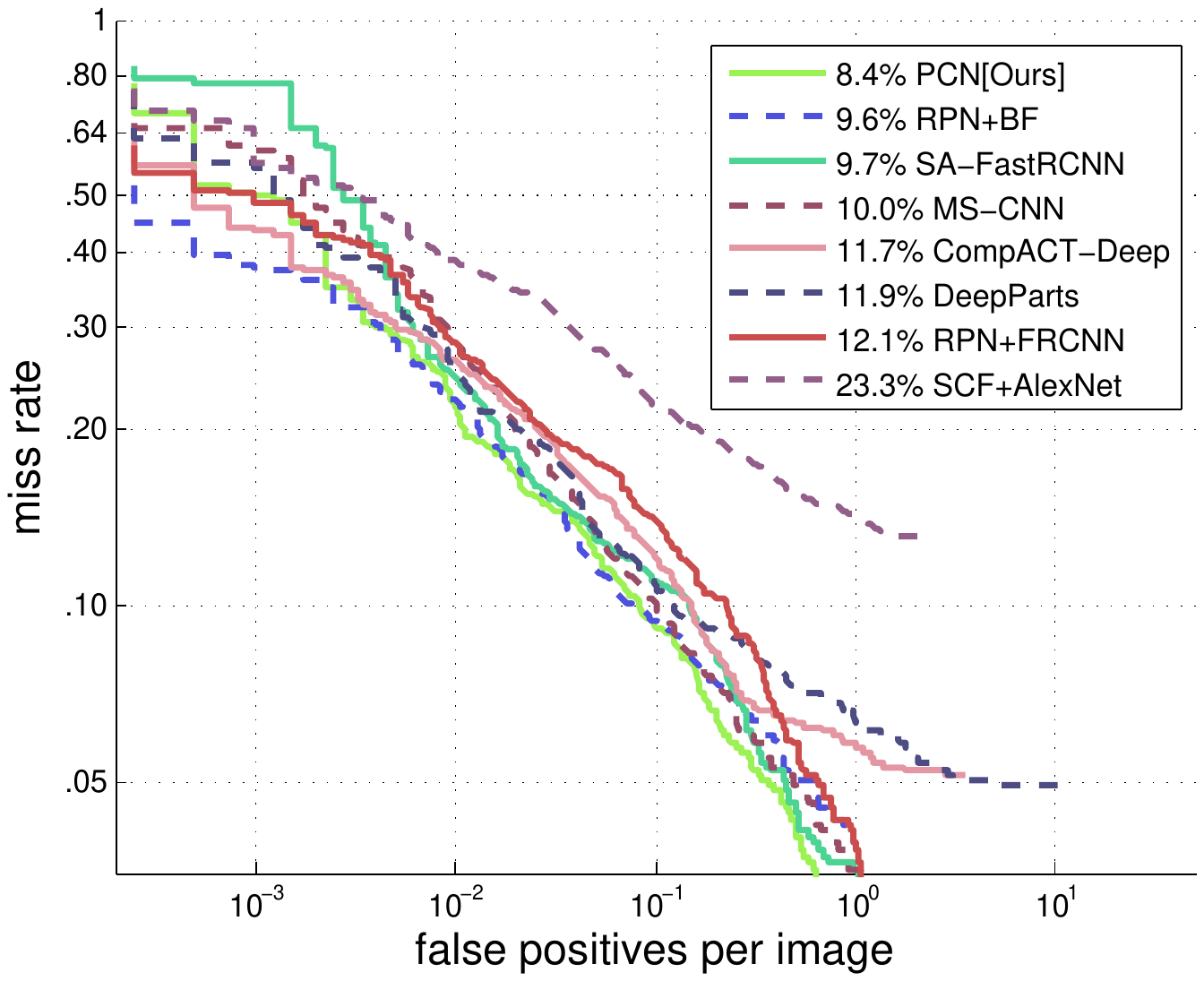}}}&
\bmvaHangBox{\fbox{\includegraphics[width=5.6cm]{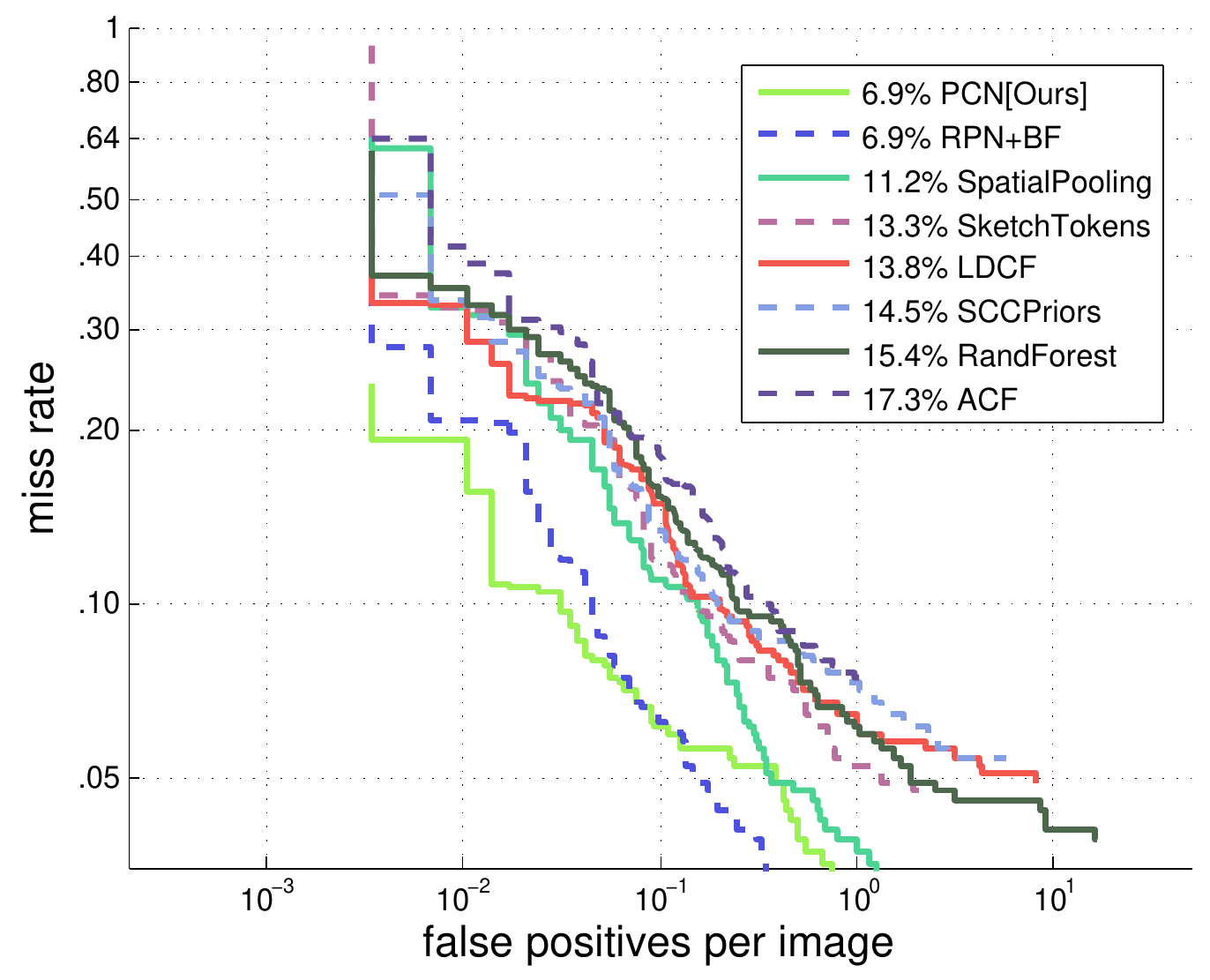}}}\\
(a)&(b)
\end{tabular}
\caption{The results on two Datasets under Reasonable evaluation (legends indicate MR) (a) Comparisons on the \textbf{Caltech} set; (b) Comparisons on the \textbf{INRIA} set.}
\label{fig:reasonable results}
\end{figure}

\begin{figure}
\centering
\includegraphics[width=10.6cm]{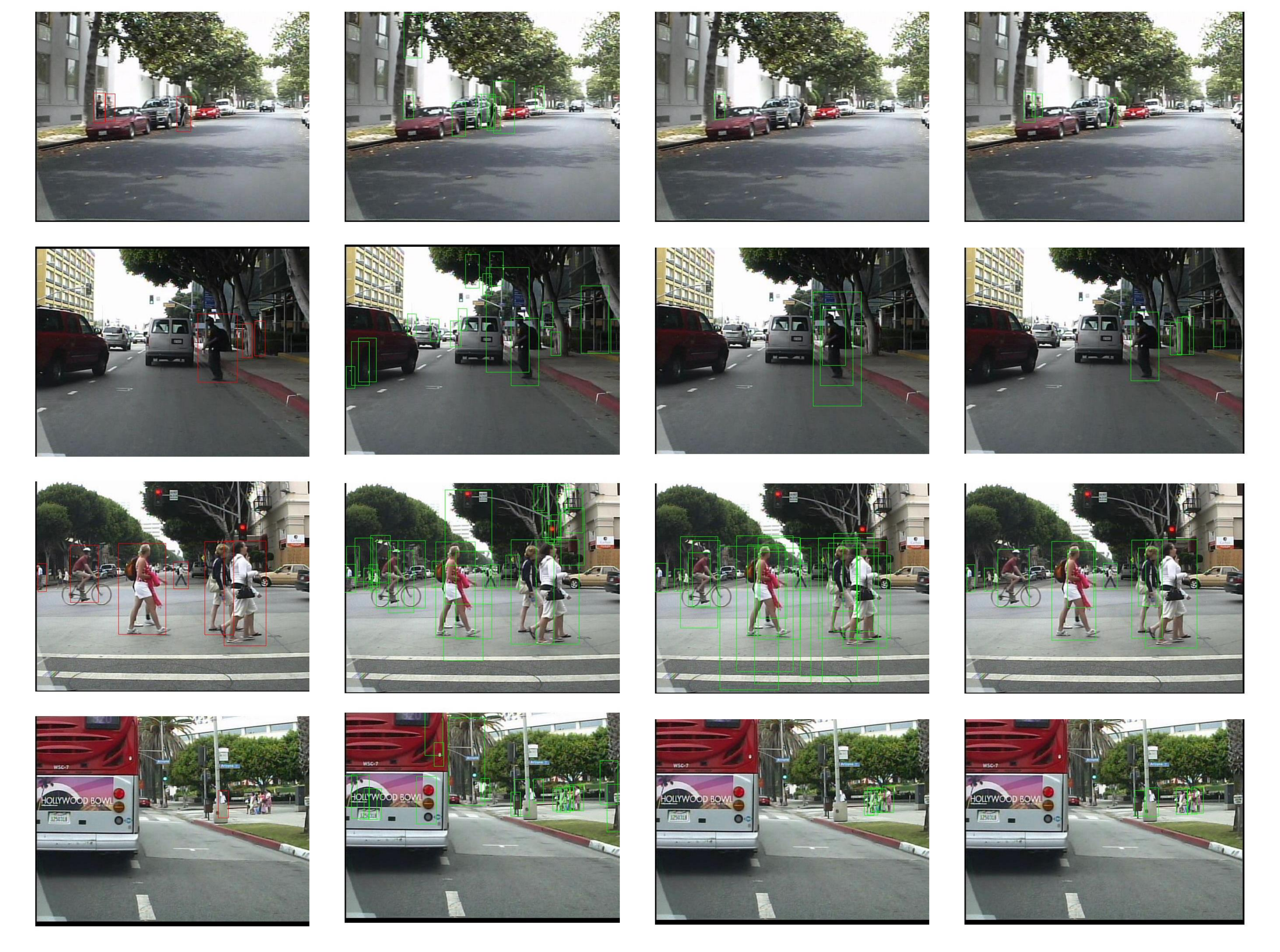}
\caption{Comparison of pedestrian detection results with other state-of-the-art methods. The first column shows the input images with ground-truths annotated with red rectangles. The rest columns show the detection results (green rectangles) of DeepParts~\cite{Tian2015Deep}, RPN+BF~\cite{Zhang2016Is} and PCN respectively. Our PCN can successfully detect more partial or heavy occluded instances with better localization accuracy compared to the other two state-of-the-art methods.}
\label{caltech result vis}
\end{figure}

\noindent \textbf{Caltech}.
Fig.~\ref{fig:reasonable results}. (a) and Table~\ref{caltech result table} show the results on Caltech (lower is better).
We compare our framework with several other state-of-the-art approaches~\cite{Hosang2015Taking,Tian2015Deep,Cai2015Learning,
li2015scale,cai2016unified,Zhang2016Is,Ren2016Faster}. Compared with these approaches, our PCN obtained a miss rate of 8.4$\%$ on reasonable setting, which is over 1.2 points better than the closest competitor (9.6$\%$ of RPN+BF~\cite{Zhang2016Is}). In the partial and heavy occlusion settings, our PCN achieved at least 3 points improvements than DeepParts~\cite{Tian2015Deep} which is specially designed for occlusion, demonstrating the effectiveness of our PCN model for occlusion handling. Visualization of detection results by state-of-the-art methods and our PCN model are shown in Fig.~\ref{caltech result vis}.

\noindent \textbf{INRIA}.
Fig.~\ref{fig:reasonable results}. (b) shows the detection results on the INRIA datasets. Our PCN model achieved a miss rate of 6.9$\%$, comparable with the best available competitor RPN+BF~\cite{Zhang2016Is}.

\subsection{Ablation Experiments}
In order to figure out which component is working, we carried out the following ablation experiments further.

\noindent \textbf{A. Part Branch}
In this subsection, we explore the effect of part branch.
In PCN, part branch divides the pedestrian box into several parts. Each part corresponds to multiple part detector (RoI-Pooling from different layers).
LSTM is introduced to encode the coherence between parts and make semantic communication among them, which is crucial for occluded pedestrian detection.
To verify this, we first added part detectors on the parameter optimized basic model RPN+RFCNN$_{opt}$, the detector scores of all parts were averaged for the final score of part branch. As we can see, Base+Part$_{avg}$ has better performance than the Base method, especially the performance of Occ.partial (24.0$\%$ vs 21.0$\%$). However, the performance of Occ.heavy increased a little, which indicates that simply splitting pedestrian box to body parts to design part detector is not enough to recognize heavy occluded pedestrian. When introducing LSTM for encoding the coherence between parts (Base+Part+LSTM), the performance of Occ.partial and Occ.heavy improved from 21.0$\%$/64.6$\%$ to 17.0$\%$/58.9$\%$, respectively, showing that the communication of semantic parts is very important for pedestrian detection, especially for these heavy occluded instances.

\noindent \textbf{B. Context Branch}
In this subsection, we explore the effect of context branch. 
To reveal the importance of context information in pedestrian detection, we added a context branch based on the basic model. One scale context only uses the context information generated by window size of WS$\times$HS, Table~\ref{caltech ablations table} shows the performance comparison when choosing different context scales (Base+Context(S=x)), obtaining a MR of 9.8$\%$/9.9$\%$/9.6$\%$ for x=1.5/1.8/2.1, respectively. This demonstrated that context plays a key role in pedestrian detection. Interesting, the performance using an IoU threshold of 0.75 (Over.75) increased a lot, the improvement of localization accuracy may be caused by the context information and extra bounding box regulator. For adaptive context selection, we introduced Maxout to integrate context scale. Base+Context (Maxout) achieved a MR of 9.0$\%$, better than those models with single context scale. Because of length limited, we just listed 3 scales (x=1.5, 1.8, 2.1) for illustration.

\begin{table}
\footnotesize
  \centering
  \begin{tabular}{p{2.8cm}|c|c|c|c|c|c}
    \toprule[2pt]
    Methods & Reasonable & All & Occ.none & Occ.partial & Occ.heavy & Over.75  \\ \toprule[1pt]	\toprule[1pt]
    RPN+FRCNN$_{opt}$(Base)	& 12.1  & 65.4  & 10.5 & 24.0 & 64.9 & 48.5 \\
    \toprule[1pt]
    Base+Part$_{avg}$ & 10.1 & 62.8 & 8.1 & 21.0 & 64.6 & 39.0 \\
    Base+Part+LSTM  & 9.4 & 62.6 & 8.1 & 17.0 & 58.9  & 38.5 \\

    \toprule[1pt]
    Base+Context(S=1.5)  & 9.8 & 63.5 & 8.1 & 19.4 & 60.1 & 38.0 \\
    Base+Context(S=1.8) & 9.9 & 63.4 & 8.3 & 20.6 & 59.8 & 36.5 \\
    Base+Context(S=2.1)  & 9.6 & 63.5 & 8.2 & 17.8 & 59.9 & 36.1 \\
    Base+Context(Maxout)  & 9.0 & 62.6 & 7.5 & 17.5 & 57.5 & 34.6 \\
    \toprule[1pt]
    PCN(full) & 8.4 & 61.8 & 7.0 & 16.4 & 56.7 & 34.8  \\
    \toprule[2pt]
  \end{tabular}
  \caption{Detailed breakdown performance comparisons of ablation experiments: the effect of part branch and context branch. RPN+FRCNN$_{opt}$: the parameter optimized RPN+FRCNN; Base+Part$_{avg}$: basic model add part detectors; Base+Part+LSTM: basic model add part detectors and semantic parts communication; Base+Context(S=x): basic model add context with single scale x (x=1.5,1.8,2.1); Base+Context(Maxout): basic model add context using Maxout for adaptively scale selection.}
  \label{caltech ablations table}
\end{table}

\section{Conclusion}
In this paper, we proposed the part and context network (PCN). PCN specially utilizes two sub-networks which detect the pedestrians through body parts semantic information and context information, respectively. Extensive experiments demonstrated that the proposed PCN is superior in detecting occluded pedestrian instances and achieving better localization accuracy.

\section*{Acknowledgement}
This work was supported by the National Natural Science Foundation of China (61671125, 61201271, 61301269), and the State Key Laboratory of Synthetical Automation for Process Industries (NO. PAL-N201401).

\bibliography{egbib}

\begin{thebibliography}{32}
\providecommand{\natexlab}[1]{#1}
\providecommand{\url}[1]{\texttt{#1}}
\expandafter\ifx\csname urlstyle\endcsname\relax
  \providecommand{\doi}[1]{doi: #1}\else
  \providecommand{\doi}{doi: \begingroup \urlstyle{rm}\Url}\fi

\bibitem[Appel et~al.(2013)Appel, Fuchs, Doll{\'a}r, and
  Perona]{appel2013quickly}
Ron Appel, Thomas~J Fuchs, Piotr Doll{\'a}r, and Pietro Perona.
\newblock Quickly boosting decision trees-pruning underachieving features
  early.
\newblock In \emph{ICML (3)}, pages 594--602, 2013.

\bibitem[Bell et~al.(2016)Bell, Lawrence~Zitnick, Bala, and
  Girshick]{Bell2016Inside}
Sean Bell, C~Lawrence~Zitnick, Kavita Bala, and Ross Girshick.
\newblock Inside-outside net: Detecting objects in context with skip pooling
  and recurrent neural networks.
\newblock In \emph{Proceedings of the IEEE Conference on Computer Vision and
  Pattern Recognition}, pages 2874--2883, 2016.

\bibitem[Bilal et~al.(2016)Bilal, Khan, Khan, and Kyung]{bilal2016low}
Muhammad Bilal, Asim Khan, Muhammad Umar~Karim Khan, and Chong-Min Kyung.
\newblock A low complexity pedestrian detection framework for smart video
  surveillance systems.
\newblock \emph{IEEE Transactions on Circuits and Systems for Video
  Technology}, 2016.

\bibitem[Broggi et~al.(2016)Broggi, Zelinsky, {\"O}zg{\"u}ner, and
  Laugier]{broggi2016intelligent}
Alberto Broggi, Alex Zelinsky, {\"U}mit {\"O}zg{\"u}ner, and Christian Laugier.
\newblock Intelligent vehicles.
\newblock In \emph{Springer Handbook of Robotics}, pages 1627--1656. Springer,
  2016.

\bibitem[Cai et~al.(2015)Cai, Saberian, and Vasconcelos]{Cai2015Learning}
Zhaowei Cai, Mohammad Saberian, and Nuno Vasconcelos.
\newblock Learning complexity-aware cascades for deep pedestrian detection.
\newblock In \emph{Proceedings of the IEEE International Conference on Computer
  Vision}, pages 3361--3369, 2015.

\bibitem[Cai et~al.(2016)Cai, Fan, Feris, and Vasconcelos]{cai2016unified}
Zhaowei Cai, Quanfu Fan, Rogerio~S Feris, and Nuno Vasconcelos.
\newblock A unified multi-scale deep convolutional neural network for fast
  object detection.
\newblock In \emph{European Conference on Computer Vision}, pages 354--370.
  Springer, 2016.

\bibitem[Chen et~al.(2014)Chen, Papandreou, Kokkinos, Murphy, and
  Yuille]{Chen2014Semantic}
Liang-Chieh Chen, George Papandreou, Iasonas Kokkinos, Kevin Murphy, and Alan~L
  Yuille.
\newblock Semantic image segmentation with deep convolutional nets and fully
  connected crfs.
\newblock \emph{arXiv preprint arXiv:1412.7062}, 2014.

\bibitem[Dalal and Triggs(2005)]{Dalal2005Histograms}
Navneet Dalal and Bill Triggs.
\newblock Histograms of oriented gradients for human detection.
\newblock In \emph{IEEE Computer Society Conference on Computer Vision and
  Pattern Recognition}, pages 886--893, 2005.

\bibitem[Dollš¢r et~al.(2012)Dollš¢r, Wojek, Schiele, and
  Perona]{Doll2012Pedestrian}
P~Dollš¢r, C~Wojek, B~Schiele, and P~Perona.
\newblock Pedestrian detection: an evaluation of the state of the art.
\newblock \emph{IEEE Transactions on Pattern Analysis and Machine
  Intelligence}, 34\penalty0 (4):\penalty0 743--761, 2012.

\bibitem[Du et~al.(2016)Du, El{-}Khamy, Lee, and
  Davis]{DBLP:journals/corr/DuELD16}
Xianzhi Du, Mostafa El{-}Khamy, Jungwon Lee, and Larry~S. Davis.
\newblock Fused {DNN:} {A} deep neural network fusion approach to fast and
  robust pedestrian detection.
\newblock \emph{CoRR}, abs/1610.03466, 2016.
\newblock URL \url{http://arxiv.org/abs/1610.03466}.

\bibitem[Gidaris and Komodakis(2015)]{Gidaris2015Object}
Spyros Gidaris and Nikos Komodakis.
\newblock Object detection via a multi-region and semantic segmentation-aware
  cnn model.
\newblock In \emph{Proceedings of the IEEE International Conference on Computer
  Vision}, pages 1134--1142, 2015.

\bibitem[Girshick(2015)]{Girshick2015Fast}
Ross Girshick.
\newblock Fast r-cnn.
\newblock In \emph{IEEE International Conference on Computer Vision}, pages
  1440--1448, 2015.

\bibitem[Girshick et~al.(2014)Girshick, Donahue, Darrell, and
  Malik]{girshick2014rich}
Ross Girshick, Jeff Donahue, Trevor Darrell, and Jitendra Malik.
\newblock Rich feature hierarchies for accurate object detection and semantic
  segmentation.
\newblock In \emph{Proceedings of the IEEE conference on computer vision and
  pattern recognition}, pages 580--587, 2014.

\bibitem[Goodfellow et~al.(2013)Goodfellow, Warde-Farley, Mirza, Courville, and
  Bengio]{Goodfellow2013Maxout}
Ian~J Goodfellow, David Warde-Farley, Mehdi Mirza, Aaron~C Courville, and
  Yoshua Bengio.
\newblock Maxout networks.
\newblock \emph{ICML (3)}, 28:\penalty0 1319--1327, 2013.

\bibitem[Hosang et~al.(2015)Hosang, Omran, Benenson, and
  Schiele]{Hosang2015Taking}
Jan Hosang, Mohamed Omran, Rodrigo Benenson, and Bernt Schiele.
\newblock Taking a deeper look at pedestrians.
\newblock In \emph{IEEE Conference on Computer Vision and Pattern Recognition},
  pages 4073--4082, 2015.

\bibitem[Li et~al.(2015)Li, Liang, Shen, Xu, Feng, and Yan]{li2015scale}
Jianan Li, Xiaodan Liang, ShengMei Shen, Tingfa Xu, Jiashi Feng, and Shuicheng
  Yan.
\newblock Scale-aware fast r-cnn for pedestrian detection.
\newblock \emph{arXiv preprint arXiv:1510.08160}, 2015.

\bibitem[Lin et~al.(2015)Lin, Wang, Yang, and Lai]{Lin2015Discriminatively}
L.~Lin, X.~Wang, W.~Yang, and J.~H. Lai.
\newblock Discriminatively trained and-or graph models for object shape
  detection.
\newblock \emph{IEEE Transactions on Pattern Analysis and Machine
  Intelligence}, 37\penalty0 (5):\penalty0 959--72, 2015.

\bibitem[Liu et~al.(2015)Liu, Rabinovich, and Berg]{Liu2015ParseNet}
Wei Liu, Andrew Rabinovich, and Alexander~C Berg.
\newblock Parsenet: Looking wider to see better.
\newblock \emph{arXiv preprint arXiv:1506.04579}, 2015.

\bibitem[Luo et~al.(2014)Luo, Tian, Wang, and Tang]{Luo2014Switchable}
Ping Luo, Yonglong Tian, Xiaogang Wang, and Xiaoou Tang.
\newblock Switchable deep network for pedestrian detection.
\newblock In \emph{2014 IEEE Conference on Computer Vision and Pattern
  Recognition (CVPR)}, pages 899--906, 2014.

\bibitem[Mathias et~al.(2013)Mathias, Benenson, Timofte, and
  Van~Gool]{Mathias2013Handling}
Markus Mathias, Rodrigo Benenson, Radu Timofte, and Luc Van~Gool.
\newblock Handling occlusions with franken-classifiers.
\newblock In \emph{Proceedings of the IEEE International Conference on Computer
  Vision}, pages 1505--1512, 2013.

\bibitem[Mikolajczyk et~al.(2004)Mikolajczyk, Schmid, and
  Zisserman]{Mikolajczyk2004Human}
Krystian Mikolajczyk, Cordelia Schmid, and Andrew Zisserman.
\newblock Human detection based on a probabilistic assembly of robust part
  detectors.
\newblock In \emph{European Conference on Computer Vision}, pages 69--82.
  Springer, 2004.

\bibitem[Mohan et~al.(2001)Mohan, Papageorgiou, and Poggio]{Mohan2001Example}
A.~Mohan, C.~Papageorgiou, and T.~Poggio.
\newblock Example-based object detection in images by components.
\newblock \emph{IEEE Transactions on Pattern Analysis and Machine
  Intelligence}, 23\penalty0 (4):\penalty0 349--361, 2001.

\bibitem[Ouyang et~al.(2017)Ouyang, Wang, Zhu, and Wang]{Ouyang2017Learning}
Wanli Ouyang, Ku~Wang, Xin Zhu, and Xiaogang Wang.
\newblock Learning chained deep features and classifiers for cascade in object
  detection.
\newblock \emph{arXiv preprint arXiv:1702.07054}, 2017.

\bibitem[Ren et~al.(2016)Ren, He, Girshick, and Sun]{Ren2016Faster}
S.~Ren, K.~He, R~Girshick, and J.~Sun.
\newblock Faster r-cnn: Towards real-time object detection with region proposal
  networks.
\newblock \emph{IEEE Transactions on Pattern Analysis Machine Intelligence},
  pages 1--1, 2016.

\bibitem[Ren et~al.(2015)Ren, He, Girshick, Zhang, and Sun]{Ren2015Object}
Shaoqing Ren, Kaiming He, Ross Girshick, Xiangyu Zhang, and Jian Sun.
\newblock Object detection networks on convolutional feature maps.
\newblock \emph{IEEE Transactions on Pattern Analysis and Machine
  Intelligence}, 50\penalty0 (1):\penalty0 815--830, 2015.

\bibitem[Surhone et~al.(2010)Surhone, Tennoe, and Henssonow]{Surhone2010Long}
Lambert~M. Surhone, Mariam~T. Tennoe, and Susan~F. Henssonow.
\newblock Long short term memory.
\newblock \emph{Betascript Publishing}, 2010.

\bibitem[Tian et~al.(2015{\natexlab{a}})Tian, Luo, Wang, and
  Tang]{Tian2015Deep}
Yonglong Tian, Ping Luo, Xiaogang Wang, and Xiaoou Tang.
\newblock Deep learning strong parts for pedestrian detection.
\newblock In \emph{IEEE International Conference on Computer Vision}, pages
  1904--1912, 2015{\natexlab{a}}.

\bibitem[Tian et~al.(2015{\natexlab{b}})Tian, Luo, Wang, and
  Tang]{Tian2015Pedestrian}
Yonglong Tian, Ping Luo, Xiaogang Wang, and Xiaoou Tang.
\newblock Pedestrian detection aided by deep learning semantic tasks.
\newblock In \emph{IEEE Conference on Computer Vision and Pattern Recognition},
  pages 5079--5087, 2015{\natexlab{b}}.

\bibitem[Wang and Ouyang(2012)]{Wang2012A}
Xiaogang Wang and Wanli Ouyang.
\newblock A discriminative deep model for pedestrian detection with occlusion
  handling.
\newblock In \emph{IEEE Conference on Computer Vision and Pattern Recognition},
  pages 3258--3265, 2012.

\bibitem[Wojek et~al.(2011)Wojek, Walk, Roth, and Schiele]{Wojek2011Monocular}
C~Wojek, S~Walk, S~Roth, and B~Schiele.
\newblock Monocular 3d scene understanding with explicit occlusion reasoning.
\newblock In \emph{The IEEE Conference on Computer Vision and Pattern
  Recognition, CVPR 2011, Colorado Springs, Co, Usa, 20-25 June}, pages
  1993--2000, 2011.

\bibitem[Zhang et~al.(2016{\natexlab{a}})Zhang, Lin, Liang, and
  He]{Zhang2016Is}
Liliang Zhang, Liang Lin, Xiaodan Liang, and Kaiming He.
\newblock Is faster r-cnn doing well for pedestrian detection?
\newblock In \emph{European Conference on Computer Vision}, pages 443--457.
  Springer, 2016{\natexlab{a}}.

\bibitem[Zhang et~al.(2016{\natexlab{b}})Zhang, Benenson, Omran, Hosang, and
  Schiele]{Zhang2016How}
Shanshan Zhang, Rodrigo Benenson, Mohamed Omran, Jan Hosang, and Bernt Schiele.
\newblock How far are we from solving pedestrian detection?
\newblock In \emph{Proceedings of the IEEE Conference on Computer Vision and
  Pattern Recognition}, pages 1259--1267, 2016{\natexlab{b}}.

\end{thebibliography}
\end{document}